\documentclass[conference]{IEEEtran}
\IEEEoverridecommandlockouts
\usepackage{cite}
\usepackage{amsmath,amssymb,amsfonts}
\usepackage{algorithmic}
\usepackage{graphicx}
\usepackage{textcomp}
\usepackage{xcolor}
\usepackage{algorithm}
\usepackage{array}
\usepackage[caption=false,font=normalsize,labelfont=sf,textfont=sf]{subfig}
\usepackage{textcomp}
\usepackage{multirow}
\usepackage{stfloats}
\usepackage{url}
\usepackage{verbatim}
\usepackage{graphicx}
\usepackage{ulem}
\usepackage{booktabs}       
\usepackage{listings}

\def\BibTeX{{\rm B\kern-.05em{\sc i\kern-.025em b}\kern-.08em
    T\kern-.1667em\lower.7ex\hbox{E}\kern-.125emX}}
\begin{document}
\title{Multi-Agent Causal Reasoning System for Error Pattern Rule Automation in Vehicles}

\author{
\IEEEauthorblockN{Hugo Math}
\IEEEauthorblockA{
BMW Group \\
Augsburg University \\
Augsburg, Germany \\
hugo.math@bmwgroup.com
}
\and
\IEEEauthorblockN{Julian Lorenz}
\IEEEauthorblockA{
Chair for Machine Learning and Computer Vision \\
Augsburg University \\
Augsburg, Germany \\
julian.lorenz@uni-a.de
}
\and
\IEEEauthorblockN{Stefan Oelsner}
\IEEEauthorblockA{
BMW Group \\
Quality Management \\
Munich, Germany \\
stefan.oelsner@bmw.de
}
\and
\IEEEauthorblockN{Rainer Lienhart}
\IEEEauthorblockA{
Chair for Machine Learning and Computer Vision \\
Augsburg University \\
Augsburg, Germany \\
rainer.lienhart@uni-a.de
}
}

\maketitle
\begin{abstract}

Modern vehicles generate thousands of different discrete events known as \textit{Diagnostic Trouble Codes} (DTCs). Automotive manufacturers use Boolean combinations of these codes, called \textit{error patterns} (EPs), to characterize system faults and ensure vehicle safety. Yet, EP rules are still manually
handcrafted by domain experts, a process that is expensive and prone to errors as vehicle complexity grows.  
This paper introduces \textbf{CAREP} (\uline{C}ausal \uline{A}utomated \uline{R}easoning for \uline{E}rror \uline{P}atterns), a multi-agent system that automatizes the generation of EP rules from high-dimensional event sequences of DTCs. CAREP combines a causal discovery agent that identifies potential DTC–EP relations, a contextual information agent that integrates metadata and descriptions, and an orchestrator agent that synthesizes candidate boolean rules together with interpretable reasoning traces.  
Evaluation on a large-scale automotive dataset with over 29,100 unique DTCs and 474 error patterns demonstrates that CAREP can automatically and accurately discover the unknown EP rules, outperforming LLM-only baselines while providing transparent causal explanations.  
By uniting practical causal discovery and agent-based reasoning, CAREP represents a step toward fully automated fault diagnostics, enabling scalable, interpretable, and cost-efficient vehicle maintenance.


\end{abstract}


\section{Introduction}
\noindent Modern vehicles generate vast amounts of operational data, most notably through \textit{Diagnostic Trouble Codes} (DTCs) produced asynchronously by Electronic Control Units (ECUs). DTCs encode discrete fault events that summarize subsystem malfunctions (e.g., circuit fault, voltage issue). Compared to raw sensor data, they provide a compact and structured representation that is easier to analyze and model. Recent work has shown that DTCs can be effectively modeled with small, domain-specific language models \cite{tf, gpt, touvron2023llamaopenefficientfoundation} trained on millions of sequences in a self-supervised manner~\cite{math2024harnessingeventsensorydata, gpt, touvron2023llamaopenefficientfoundation}.

Automotive manufacturers monitor fleet quality data using \textit{error patterns} (EPs), they characterize a whole sequence of DTCs consisting of precise vehicle failures (e.g., battery issues, PCB fault). They are defined as Boolean rules over DTCs. For instance, 
\(
ep_1 = dtc_1 \;\&\; dtc_2 \;\&\; (! dtc_3 \;|\; dtc_4)
\)
is triggered whenever $dtc_1$ and $dtc_2$ occur together, but only in the absence of $dtc_3$ or the presence of $dtc_4$. EPs are central to vehicle safety and maintenance, but domain experts largely handcraft their rules. Manually identifying and updating the rules for hundreds of error patterns across thousands of vehicle types is costly, time-consuming, and prone to human mistakes. This motivates the need for automatized error pattern rule discovery.

At their core, EPs rules capture causal relationships, i.e., certain DTCs jointly act as causes of specific error patterns. Traditional causal discovery algorithms~\cite{cd_temporaldata_review} have made progress in low-dimensional tabular or temporal data, and more recent approaches address sequence-level causal discovery~\cite{tf_causalinterpretation_neurips_2023}. However, two challenges remain: (1) most methods ignore causal strength, preventing them from distinguishing excitatory and inhibitory effects required to form Boolean rules, (2) existing algorithms scale poorly in high-dimensional settings, where real-world datasets often contain thousands of event types\cite{hasan2023a}.

\begin{figure}[!t]
    \centering
    \includegraphics[width=0.46\textwidth]{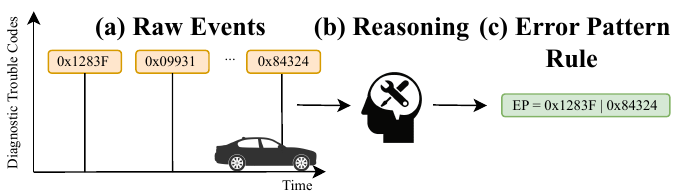} 
    \caption{\textbf{Illustration of error pattern automation in modern vehicles.} (a) A vehicle with an unknown defect generates a sequence of Diagnostic Trouble Codes (DTCs) over time. (b) A domain expert analyzes the DTCs and the metadata associated with the vehicle (model, descriptions, etc.). (c) Then, provide a Boolean rule to identify this error pattern.}
    \label{fig:illustration}
\end{figure}
Beyond causality, automation requires interpretability. Presenting unexplained rules is risky in safety-critical domains like the automotive industry or in medical health data. Large Language Models (LLMs)~\cite{OpenAI_GPT4_2023, deepseekai2025deepseekr1incentivizingreasoningcapability} have recently demonstrated strong reasoning and in-context learning capabilities~\cite{icl, zhang2024largelanguagemodelsinterpolated}, especially when using Chain-of-thought \cite{chainofthoughts} (CoT) to guide them through step-by-step thought processes.

 Paired with the growing paradigm of \textit{agentic systems}~\cite{agents, zhao2025agenticrarediseasediagnosis, biomedicalagent}, where software agents orchestrate and query LLMs, this opens the door to human-interpretable, automatized causal reasoning in complex industrial pipelines \cite{2024arXiv240715073D, le2025multiagentcausaldiscoveryusing}. However, as pointed out by \cite{mirzadeh2025gsmsymbolic}, LLMs are sensitive and fragile. For tasks requiring the selection of the correct tokens, their accuracy decreases exponentially with the number of tokens, making them unsuitable for high-dimensional event sequences.

\textbf{Contributions.} In this paper, we introduce \textbf{CAREP} (\uline{C}ausal \uline{A}utomated \uline{R}easoning for \uline{E}rror \uline{P}atterns), the first multi-agent system for automatic discovery of EP rules in high-dimensional automotive event sequences. CAREP integrates:
\begin{enumerate}
    \item a \textit{causal discovery agent} that estimates candidate causes and causal indicators for each unknown EP.
    \item a \textit{contextual information agent} that incorporates DTC descriptions, known EP rules, and vehicle metadata.
    \item an \textit{orchestrator agent} that coordinates reasoning and outputs candidate rules with natural-language traceable reasoning chains.   
\end{enumerate}

\noindent We evaluate CAREP on a real-world dataset of \(29,100\) DTCs and \(474\) known EPs rules by masking a random subset of the known rules and testing the framework’s ability to identify them. Results show that CAREP outperforms LLM-only baselines by a large margin while offering interpretable explanations at scale.

To the best of our knowledge, this is the first work that applies multi-agent systems with causal reasoning to automatize error pattern rule discovery. CAREP takes a step toward practical automation of vehicle diagnostics, with the potential to reduce maintenance costs and improve road safety.

\section{Related Work}

\subsection{Event Sequence Modeling}
\noindent Event sequences are typically represented as a series of time-stamped discrete events \(s = \{(t_1, x_1), \ldots, (t_L, x_L)\}\) where \(0 \leq t_1 < \ldots \leq t_L\) denotes the time of occurrence of event type \(x_i \in \mathbb{X}\) drawn from a finite vocabulary \(\mathbb{X}\). In multi-label settings, a binary label vector \(\boldsymbol{y} \in \{0, 1\}^{|\mathbb{Y}|}\) is attached to \(s\) and indicates the presence of multiple outcome labels chosen from \(\mathbb{Y}\) occurring at the final time step \(t_L\). Together, this results in a multi-labeled sequence \(S = (s, (\boldsymbol{y}_L, t_L))\). 

Event sequence modeling has been widely applied to predictive tasks. For instance, in the automotive domain, Diagnostic Trouble Codes (DTCs as events) are logged asynchronously over time and used to infer failures or error patterns \cite{math2024harnessingeventsensorydata} (as labels). In healthcare, electronic health records encode temporal sequences of symptoms to perform predictive tasks \cite{MedBERT, pmlr-v219-labach23a, bihealth}. A common modeling strategy  \cite{crf, cmm} separates such event types \(\mathbb{X}\) from labels \(\mathbb{Y}\).

Transformers \cite{tf, gpt, touvron2023llamaopenefficientfoundation} have emerged as the dominant architecture for sequence modeling. 
Recent work leveraged Transformers in high-dimensional event spaces for next-event and label prediction. \cite{math2024harnessingeventsensorydata} proposed a dual Transformer architecture where one model predicts the next event type (DTC), and the other predicts label occurrence (e.g, error pattern). Through this paper, we repurpose this dual architecture for causal discovery. 


\subsection{Error Pattern Rule Automation}
\noindent
The creation of EPs (Fig~\ref{fig:illustration}) emerges as the complexity of production and vehicle manufacturing drastically increased. 
To identify new defects, domain experts analyze the sequence of diagnostic trouble codes (DTCs) for each vehicle.

Intuitively, this reasoning task is not trivial. The domain experts have to use their own past observations and knowledge about specific DTCs that might cause each EP. 
That is where the Boolean operators are useful. They allow us to construct different rules to characterize precise EPs. It sometimes requires the absence of certain DTCs (\(!\) NOT) to separate overlapping EPs. Or the presence of multiple DTCs at the same time (\(\&\) AND) or a particular culprit (\(|\) OR). 

Therefore, the different operators \((\&, |, !)\) involve different statistical perspective. Intuitively, the \(!\) operator measures inhibitory strength \cite{measure_of_causal_strength_oxford}, where the likelihood of observing \(y_1\) will be lowered if we observe \(x_{10}\) (Eq.~\eqref{eq:ep_def}). On the other hand, \(\&\) and \(|\) are more associated with excitatory events that will raise the likelihood of an EP. Thus, knowing only the causes of an EP is not sufficient to elaborate a new rule. It is necessary to have indicators that measure how DTCs are related to each other and to EPs. 
Finally, rules are subject to change and are dynamically updated by a domain expert based on the new incoming data. 

As a result, automating a rule is a difficult, dynamic problem. We try to explore this reasoning task in this paper.
\subsection{Multi-label Causal Discovery}
\noindent Multi-label Causal Discovery seeks to identify causal relationships \cite{reviewmultilabellearning} between features and labels in a Bayesian Network (BN) \cite{pearl_1998_bn, optimal_feature_set_cd}.
While classical constraint-based algorithms have shown success on low-dimensional tabular data \cite{constrainct_based_cd, causality_based_feature_selection_2019}, their application to event sequences with multi-label outputs remains challenging due to: (1) \textit{dimensionality}—thousands of event types increase super-exponentially the number of graphs, (2) \textit{sparsity}—multi-hot encodings often underrepresent rare but important events, (3) \textit{distributional assumptions}—such as linearity or Gaussian noise, which rarely hold in real-world sequences \cite{cd_temporaldata_review}. 

Contemporary work suggests a divide-and-conquer approach \cite{divide_conquer_mb_discovery} that reformulates classical causal discovery for high-dimensional datasets into a graph aggregation problem. \cite{laborda_ring_based_distirbuted} introduce a ring-based distributed algorithm for learning high-dimensional BN, \cite{divide_conquer_mb_discovery, recursive_mb_approach_pmlr_large_scale} explore distributed approaches for large-scale causal structure learning.

\section{Notations}
\noindent We use capital letters (e.g., \(X\)) to denote random variables, lower-case letters (e.g., \(x\)) for their realizations, and bold capital letters (e.g., \(\boldsymbol{X}\)) for sets of variables. Let \(\boldsymbol{U}\) denote the set of all (discrete) random variables. We define the event set \(\boldsymbol{X} = \{X_1, \ldots, X_n\} \subset \boldsymbol{U}\), and the label set \(\boldsymbol{Y} = \{Y_1, \ldots, Y_n\} \subset \boldsymbol{U}\). EPs are denoted as labels \(\boldsymbol{Y}\) and DTCs as events \(\boldsymbol{X}\). Unknown EPs denote the EPs for which we didn't identify a name or a rule yet.
\section{Methodology}
\noindent Let \(\mathcal{D} = \{S^1, \cdots, S^m\}\) be a dataset of multi-labeled sequences. Each label \(y_j\) is defined as a Boolean rule between events \(\boldsymbol{X}\) (Eq.~\eqref{eq:ep_def}). Such that if in a sequence \(S^k\), the Boolean rule is True for some label \(y_j\), then it is present in \(S^k\).

\(\mathcal{D}\) contains multiple labels with unknown Boolean rules denoted as \(y^{un} \in \mathbb{Y}\). 
Our goal in this paper is to find the unknown rule of each label \(y^{un}\) using the observed sequences. Eq.~\ref{eq:ep_def} shows an EP rule for \((y_1)\) based on some diagnosis trouble codes \((x_i)\).
\begin{equation} \label{eq:ep_def}
  y_1 = x_1 \;  \& \; x_5 \;\&\; x_8 \; \& \; (x_{12} \; | \; x_3) \; \& \; !x_{10} \;\& \; !x_{20}
\end{equation}
\noindent The general architecture of \textbf{CAREP} can be seen in Fig.~\ref{fig:carep}. The agentic system outputs a reasoning explanation for each estimated error pattern rule. To increase prediction's plurality, CAREP outputs \(5\) rules per EP with different levels of confidence (high, medium, low) that are reflected in the provided explanation. 
Each agent is given a system prompt to specify their task.
We will now dive into each block separately.

\subsection{Supervised Unknown Class Learning}
\noindent As we saw earlier, domain experts rely on past knowledge and observations. From this, we would like to have a way to represent past knowledge and retrieve it efficiently during inference. Thanks to self-supervised learning \cite{NIPS1999_e6384711, touvron2023llamaopenefficientfoundation}, we can pretrain small domain-specific language models (SLMs) on a vast corpus of DTC sequences by performing next-DTC prediction.

Let's define two SLMs trained on \(\boldsymbol{D}\) using next event and label prediction \cite{math2024harnessingeventsensorydata}, respectively \(\text{Tf}_x, \text{Tf}_y\) parametrized by \(\theta_x, \theta_y\). They model the conditional probability distribution \(P_{\theta_x}(X_i|\boldsymbol{Z}), P_{\theta_y}(Y_j|\boldsymbol{Z})\), where \(\boldsymbol{Z} = (x_1, \cdots, x_{i-1}) = S_{< i}\) are the past events such that: 
\begin{align}
\text{Tf}_x(S_{< i}) &= \textit{Softmax}(\boldsymbol{h}^{x}_{i-1}) = P_{\theta_x}(X_i| \boldsymbol{Z}) \label{eq:prob_x} \\
\text{Tf}_y(S_{\leq i}) &= \textit{Sigmoid}(\boldsymbol{h}^{y}_{i}) = P_{\theta_y}(Y|X_i, \boldsymbol{Z})\label{eq:prob_y}
\end{align}

Here, \(\boldsymbol{h}^{x}_{i-1}, \boldsymbol{h}^{y}_{i} \in \mathbb{R}^{d}\) are the logits produced by the two Transformer heads of \( \text{Tf}_x\) and \(\text{Tf}_y\) parametrized by \(\theta_x, \theta_y\). \(X_i\) is the event at step \(i\) in \(S\). The majority of \(\text{Tf}_x\) (except the heads) serves as a backbone for \(\text{Tf}_y\).

We explicitly model the labels \(y^{un}\) as the target in the autoregressive classifier \(\text{Tf}_y\)'s output logits. This forms a supervised unknown class learning strategy and enables us to extract the posteriors \(P_{\theta_y}(Y = y^{un}|\boldsymbol{Z}=z_i)\) and \(P_{\theta_y}(Y = y^{un}|X_i=x_i, \boldsymbol{Z}=z_i)\). In other words, we can predict the next EPs based on the past DTCs.

\subsection{One-Shot Causal Discovery}\label{sec:cd}
\noindent To provide new causes for unknown error patterns, we use the learned conditionals \(P_{\theta_y}(Y_j|\boldsymbol{Z})\) and \(P_{\theta_y}(Y_j|X_i, \boldsymbol{Z})\) from \(\mathcal{D}\). Thus, we must recover the causes (i.e., the DTCs noted as \(x_i\)) of a specific label \(y_j\) (i.e., an error pattern). We base our approach on causal discovery in single streams \cite{math2025oneshot, math2025towards}.

We model the random variable \(Y_j \sim Ber(\boldsymbol{p}_j)\) such that if \(Y_j = 1\), the EP \(y_j\) occurred in the sequence \(S\). \(\boldsymbol{p} \in [0, 1]^L\) is a step dependent parameter vector defined by: \[p_{j,i+1} \triangleq P_{\theta_y}(Y = y_j|X_i, \boldsymbol{Z}) = P(Y_j=1|X_i, \boldsymbol{Z})\]
Hence, we would like to assess how much additional information event \(X_i\) occurring at step \(i\) provides about label \(Y_j\) when we already know the past sequence of events \(\boldsymbol{Z} = S_{<i}\). We essentially try to answer if:
\begin{align*}
  & P(Y_{j}|X_i, \boldsymbol{Z}) = P(Y_{j}|\boldsymbol{Z}) \\
  &\Leftrightarrow D_{KL}(P(Y_{j}|X_i, \boldsymbol{Z})\|P(Y_{j}|\boldsymbol{Z})) = 0  
\end{align*}
where \(D_{KL}\) denotes the \textit{Kullback-Leibler divergence} \cite{cover1999elements}. The distributional difference between the conditionals \(P(Y_j|X_i, \boldsymbol{Z}), P(Y_j| \boldsymbol{Z})\) is akin to Information Gain \(I_G\) \cite{quinlan:induction} conditioned on past events:
{\small
\begin{equation}\label{eq:info_gain}
    I_G(Y_j, x_i|z_i) \triangleq D_{KL}(P(Y_j|X_i=x_i, \boldsymbol{Z} = z_i)) || P(Y_j|\boldsymbol{Z}=z_i)) 
\end{equation}
}

\noindent Which is equal to the difference between the conditional entropies \cite{cover1999elements} denoted as \(H\): 
\begin{equation}
    I_G(Y_j, x_i|z_i) = H(Y_j|z_i) - H(Y_j|x_i, z_i)
\end{equation}

\noindent More generally, we can access the conditional independence of event \(X_i\) and label \(Y_{j}\) using the conditional mutual information (CMI) \cite{cover1999elements} which is simply the expected value over \(x_i, z_i\) of the information gain \(I_G(Y_j, x_i|z_i)\) such as:
\begin{align}
    I(Y_{j}, X_i|\boldsymbol{Z}) &\triangleq H(Y_j|\boldsymbol{Z}) - H(Y_j|\boldsymbol{Z}, X_i) \\
    &= \mathbb{E}_{x_i, z_i}[I_{G}(Y_j, X_i=x_i|\boldsymbol{Z}=z_i)])\label{eq:cmi_theorique}
\end{align}

\noindent If \(I(Y_j, X_i|\boldsymbol{Z})\) is greater than \(0\) \cite{cover1999elements}, we say that \(X_i\) is a cause of \(Y_j\) and append it to the list of potential causes of label \(Y_j\). In practice, causality is determined using a label-specific threshold \(\theta_j\), defined as \(\theta_j=\mu_{Y_j} + \gamma \sigma_{Y_j},\) where \(\mu_{Y_j}\) and \(\sigma_{Y_j}\) denote the mean and standard deviation of the CMI values over \(S^k\), and \(\gamma\) is a fixed sensitivity parameter.


The expectation in Eq.~\( \eqref{eq:cmi_theorique}\) is computed using a Monte-Carlo simulation by sampling \(N\) similar context \(\boldsymbol{Z}\) from \(\text{Tf}_x\). 
Such that for each position \(i\) in the sequence, we generate \(N\) plausible next tokens by using a combination of top-k and nucleus sampling \cite{Holtzman2020The}. 

To ensure stable conditional entropy estimates and reliable predictions from \(\text{Tf}_y\), the CMI is computed after observing \(c\) events (\textit{context}) \cite{math2024harnessingeventsensorydata}. This design choice enables out-of-the-box parallelization since the CMI estimations are independently performed for all positions \(i \in [c, L]\). This transitions the time complexity from \(\mathcal{O}(\text{BS} \times N \times L)\) to \(\mathcal{O}(1)\) per batch.

An ablation of the quality of the pretraining of \(\text{Tf}_x, \text{Tf}_y\) on the classification test set used in the experiment is given in the Appendix. 
Finally, this step generates \(m\) graphs \(\{\mathbb{G}^k\}^m_{k=1}\) in parallel from the dataset \(\mathcal{D}\) of \(m\) sequences. We denote it as the \emph{one-shot causal discovery} phase.
 
\subsection{Causal Indicators}\label{sec:causal_indicators}
\noindent Given that we can estimate conditional distributions, we define the \textit{causal indicator} \(\mathcal{C} \in [-1, 1]\) \cite{measure_of_causal_strength_oxford} between an event \(X_i\) and a label \(Y_j\) under context \(\boldsymbol{Z}\) that we assume fixed for every measurement \cite{measure_of_causal_strength_oxford} as the Average Causal Effect of event \(X_i\) on \(Y_j\) (ACE or Eells measure \cite{Eells_1991}) such as:
\begin{equation}\label{eq:causal_indic}
\mathcal{C}(Y_j, X_i) := \mathbb{E}_{Z}[P(Y_j \mid X_i, Z) - P(Y_j \mid Z)]
\end{equation}
This enables an easier interpretation, for instance, if \(\mathcal{C} < 0\), then \(X_i\) inhibits the occurrence of \(Y_j\). Otherwise if \(\mathcal{C} > 0,\) the cause \(X_i\) raises the probability of its effect \cite{Eells_1991}. We employ the term causal \textit{indicator} to separate from causal strength measures, which, if using this formulation, can be problematic as pointed out by \cite{quantifyingcausalinfluence}. Here, it serves more as an indication to the causal estimator agent. 

\begin{figure*}[!t]
    \centering
    \includegraphics[width=0.9\textwidth]{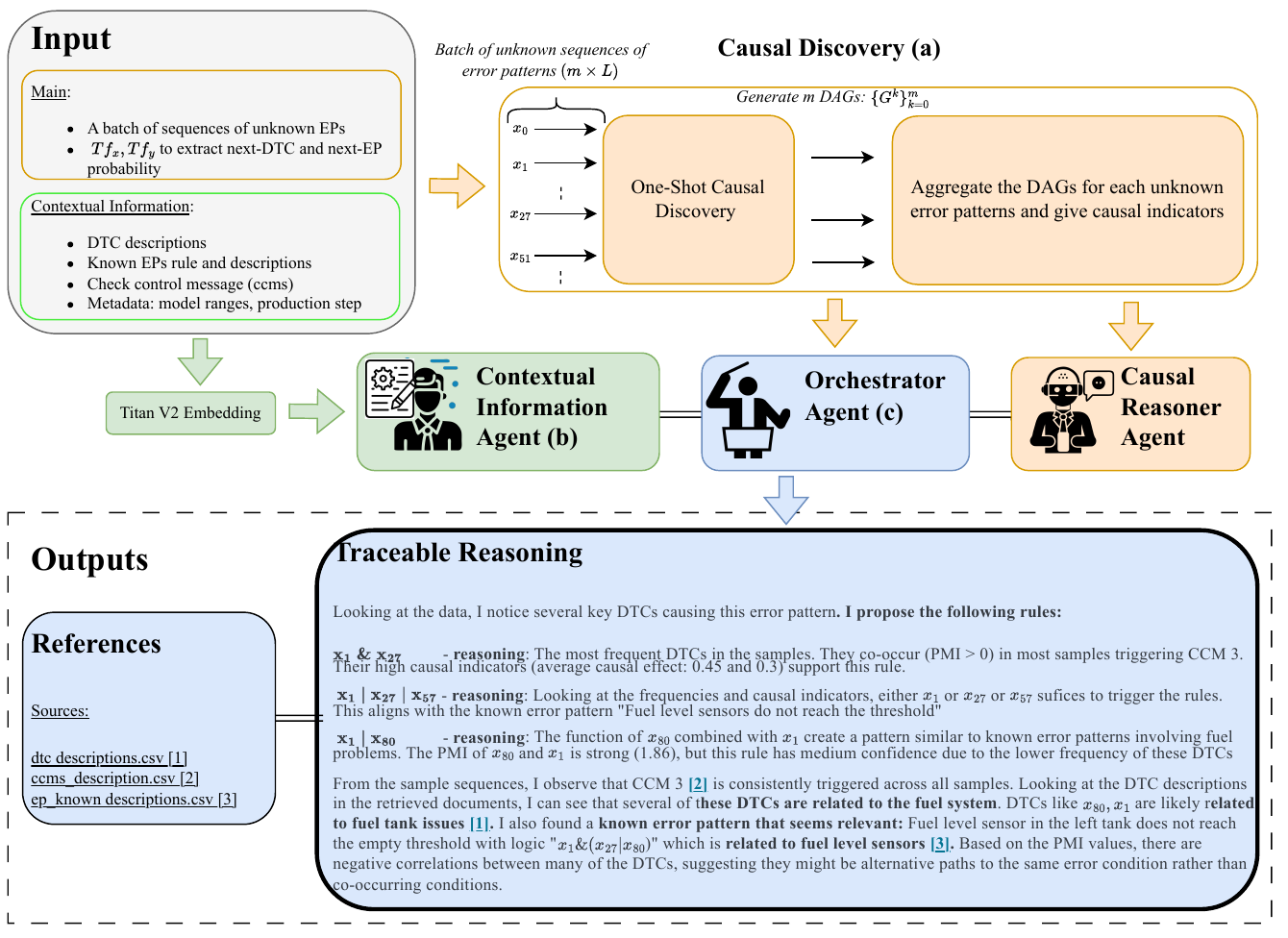} 
    \caption{\textbf{CAREP: A causal reasoning agentic system for error pattern automation}. (a) Represents the causal discovery phase where we extract candidates DTC causes for the unknown error pattern \(Y_1\), with their causal indicators. It then feeds the causal reasoner agent. (b) The descriptions and metadata are extracted through a Titan V2 Embeddings and fed into the contextual information agent. (c) The orchestrator agent manages the two agents and provides a traceable reasoning to explain why the unknown error pattern potentially has these rules. Traceable reasoning is truncated to 3 rules.} 
    \label{fig:carep}
\end{figure*}

\subsection{Adaptive Threshold for Rule Aggregation}
\noindent The causal discovery step produces one-shot directed acyclic graphs (DAGs) at the level of individual sequences $S^k$. 
To obtain a global rule for each label $Y_j$, these graphs must be aggregated. We employ a tailored majority voting to obtain global graphs \cite{survey_bn_struct_learning, consensus_bn_freq_cutoff_no}.

For each candidate edge $X_i \to Y_j$, we compute the empirical frequency 
$\hat{\pi}_{i,j}$, i.e., the fraction of one-shot DAGs for which the edge appears. 
This aggregation turns a collection of noisy, instance-specific graphs into a frequency-based edge distribution for each label.

A key difficulty arises from the long-tail distribution of labels \cite{longtailissue}. 
For head labels (large support $m_j$), empirical frequencies $\hat{\pi}_{i,j}$ are 
reliable estimates; for tail labels (small $m_j$), they are highly variable and 
susceptible to noise. Using a fixed threshold across labels would either suppress 
valid edges in common labels or include spurious edges in rare ones.

To address this, we introduce a label-specific adaptive threshold $\tau_j$ applied 
to the empirical edge frequencies $\hat{\pi}_{i,j}$. 
Rare labels require higher thresholds to regularize noisy edges, while common labels 
can afford lower thresholds to retain weaker but genuine causal relations. 
We parameterize $\tau_j$ as a logistic decay function of the support $m_j$:
\begin{equation}
\tau_j(m_j)= (\tau_{\max} - \tau_{\min}) \cdot \frac{1}{1 + e^{\alpha(\log m_j - \log m_0)}} + \tau_{\min}
\label{eq:adaptive_threshold}
\end{equation}
where \(\tau_{\max}=0.9, \tau_{\min}=0.05\), \(m_0\) is set to the median of all label supports and \(\alpha\) is set inversely proportional to the log-interquartile
range of supports: $\alpha= \tfrac{2 \log 3}{\log q_{75} - \log q_{25}}$. 
This ensures the transition adapts to the specific shape of the long-tail distribution.

Finally, applying these thresholds yields one aggregated DAG $\mathbb{G}^*_j$ for each label, 
providing a robust set of candidate causes tailored to both frequent and rare outcomes.

\subsection{Co-occurrence}\label{sec:co_occurence}
\noindent The different Boolean operators in the defined EPs rules (Eq.~\eqref{eq:ep_def}) imply different statistical perspectives. In the previous section, we captured the elements in the rule, i.e., the set of DTC causes for a given label. However, we should also provide the orchestrator agent with the co-occurrence estimates of event pairs to properly address the OR, AND, NOT operators.

Traditionally, Point-wise Mutual Information (PMI) \cite{cover1999elements} is used to measure co-occurrence strength. It is defined as:
\[pmi(x_i, x_j) = \log{\frac{p(x_i,x_j)}{p(x_i)p(x_j)}}\]

\noindent Along with the previous empirical frequency of event \(X_i\) in the set of causes of label \(Y_j\), denoted as \(\hat{\pi}_{i,j}\), we also compute empirical joint frequencies of pairs of events within the same label-specific graph \(\mathbb{G}^*_j\). 
Formally, given a collection of DAGs \(\{(\boldsymbol{V}_k, E_k)\}_{k=1}^m\), the marginal and joint empirical probabilities are estimated as
\[
\hat{p}(x_i \mid y_j) = \frac{1}{m_j} \sum_{k=1}^m \mathbf{1}\{x_i \in \boldsymbol{V}_k, \, y_j \in \boldsymbol{V}_k\},
\]
\[
\hat{p}(x_i, x_\ell \mid y_j) = \frac{1}{m_j} \sum_{k=1}^m \mathbf{1}\{x_i \in \boldsymbol{V}_k, \, x_\ell \in \boldsymbol{V}_k, \, y_j \in \boldsymbol{Y}_k\},
\]
where \(m_j\) denotes the number of samples in \(\mathcal{D}\) where label \(y_j\) is present, and \(\mathbf{1}\{\cdot\}\) is the indicator function.

\noindent The PMI between two events \(x_i\) and \(x_\ell\) conditioned on a label \(y_j\) is then given by
\[
\widehat{pmi}(x_i, x_\ell \mid y_j) 
= \log \frac{\hat{p}(x_i, x_\ell \mid y_j)}{\hat{p}(x_i \mid y_j)\,\hat{p}(x_\ell \mid y_j)}.
\]

\noindent Intuitively, \(\widehat{pmi}(x_i, x_\ell \mid y_j) > 0\) indicates that events \(x_i\) and \(x_\ell\) co-occur more often than expected under independence, while a negative value suggests mutual exclusivity. 

\subsection{Contextual Informations}
\noindent We add to each explanation of \(y^{un}\) contextual information about the description of DTCs, EPs, and the known error pattern rules. These descriptions are processed into embedding vectors using a \textit{Titan V2 Embedding}\footnote{https://aws.amazon.com/de/blogs/aws/amazon-titan-text-v2-now-available-in-amazon-bedrock-optimized-for-improving-rag/} from AWS Bedrock. They are then used as a Retrieval-Augmented Generation (RAG) system where the contextual information agent queries are matched to the description embeddings generated by the embedding model. This gets pulled into the context of the contextual information and orchestrator agents. 

Now, directly in the prompt (\textit{in-context learning} \cite{icl}), we randomly inject 10 sequences of DTCs carrying the same unknown EP that we want to identify. As well as the metadata, such as the model range of the vehicle and the triggered message printed on board (check control message: CCM). The overall data inputs can be seen in Fig.~\ref{fig:carep} and the JSON input given to the orchestrator in Fig.~\ref{verbatim:output_json}.

\section{Experiments}
\subsection{Settings}\label{sec:settings}
\noindent We used a \(g4dn.12xlarge\) instance from AWS Sagemaker to run comparisons. It contains 48 vCPUs and 4 NVIDIA T4 GPUs. We used a combination of F1-Score, Precision, and Recall with different averaging \cite{reviewmultilabellearning} to perform comparisons.

\subsection{Vehicle Dataset}
\noindent We evaluated our method on a real-world vehicular test set of \(m=300,000\) sequences, with \(|\mathbb{Y}|=474\) different error patterns and \(|\mathbb{X}| = 29,100\) different DTCs forming sequences of \( \approx 100 \pm 35\) events. The language models \(\text{Tf}_x\) and \(\text{Tf}_y\) were used with respectively 90M and 15M parameters \cite{math2024harnessingeventsensorydata}, they didn't see the test set during training. 
We created 5 folds of \(50,000\) sequences and randomly masked \(20\) different EPs with at least \(n \geq 100\) sequences. We set their masked rule as the ground truth for each label \(y^{un}_j\) and average the results across the 5 folds.
\subsection{Comparison}
We compared CAREP against multiple LLMs, such as Claude Sonnet 3.5 and 3.7\footnote{https://www.anthropic.com/news/claude-3-7-sonnet}, GPT4.1 and GPT4.1 mini\footnote{https://openai.com/index/gpt-4-1/}. We also wanted to see if using smaller LLMs \cite{belcak2025smalllanguagemodelsfuture} with fewer parameters impacted the evaluation. 
The LLMs are compared by using only the observed DTCs sequences and the descriptions, thus in Fig~\ref{fig:carep}, the orange part is entirely removed.
We also added CAREP without the causal indicators (Sections  \ref{sec:causal_indicators}, \ref{sec:co_occurence}). 
\begin{figure*}[!t]
    \centering
    \includegraphics[width=0.91\textwidth]{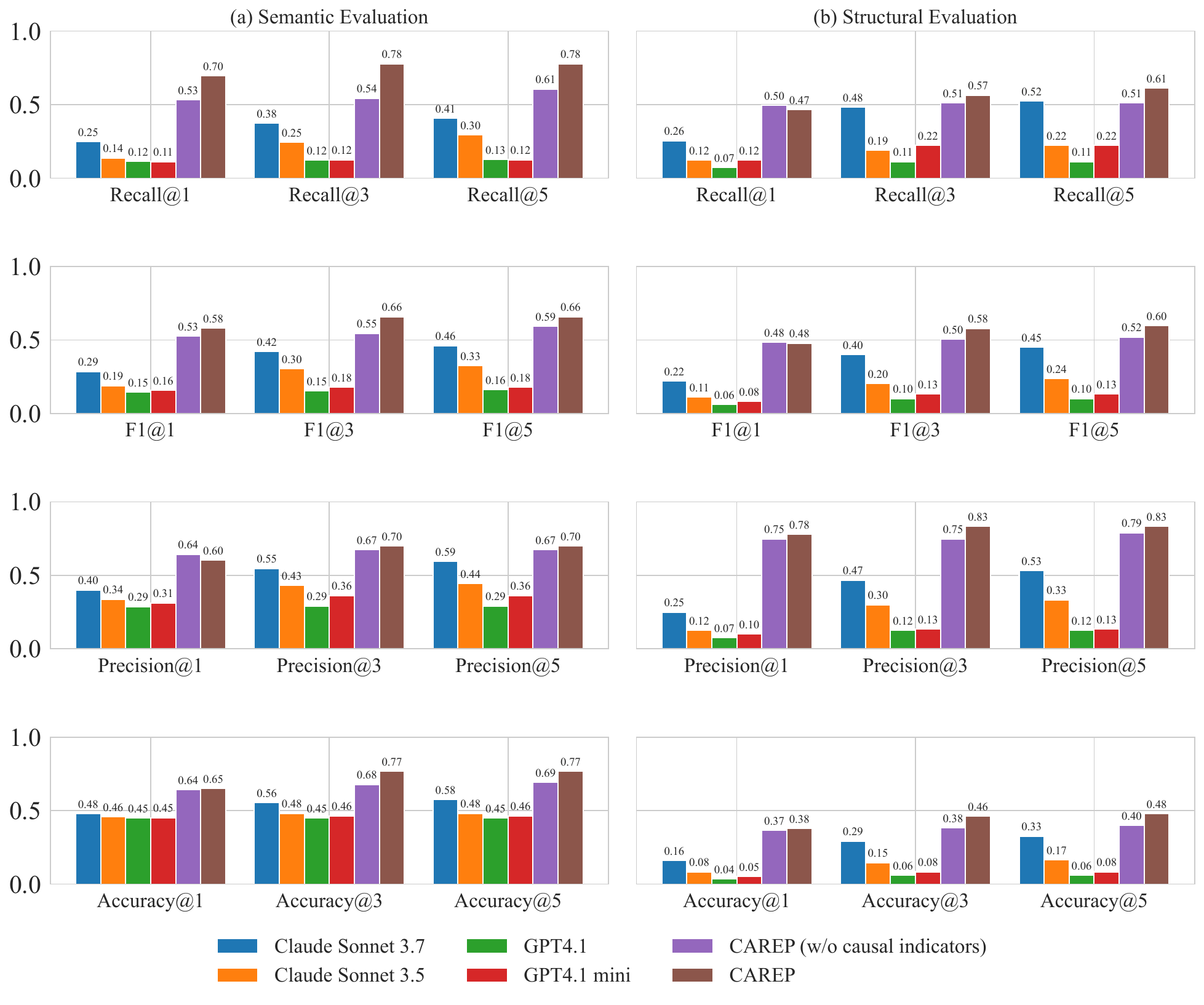} 
    \caption{Evaluation of CAREP against standalone LLMs. (a) Semantic evaluation represents how well the estimated error pattern rules fit the ground truth in terms of Boolean expression. (b) Structural evaluation or multi-label classification reveals if the estimated rules contain the correct DTCs.}
    \label{fig:carep_eval_grid}
\end{figure*}

\subsection{Evaluation}
\noindent The task of identifying error pattern rules is complex and requires multiple levels of evaluation.
Here, we distinguish between two evaluation types: (1) \textit{structural}, where we compare the Boolean rules directly as a classification set, and (2) \textit{semantic}, where we compare the corresponding truth-tables of the estimated Boolean expressions to the truth-table of the ground truth using the Sympy package \cite{sympy}.

Specifically, for (1) we evaluate: \textit{Are the DTCs in the estimated rules correct?}
For this, we divide the estimated sets and ground truth using the Boolean operators as separators and perform a standard multi-label classification:
    \[\text{dtc1 \& dtc2 \& !dtc5} \;|\; \text{dtc3} \Longrightarrow [dtc1, dtc2, dtc3, dtc5]\]


\noindent For (2) we enumerate all possible assignments of the present Boolean variables and compute the truth table of the estimated rules and the ground truth to express: \textit{Is the rule logically correct ?} We calculate the accuracy, precision, recall, f1 of the predicted value of the truth tables (e.g., Tab~\ref{tab:truth_table_example} in Appendix).

We compute the top-\(3\) and top-\(5\) metrics by using the 5 estimations from the decreasing order of confidence \textit{[HIGH, MEDIUM, LOW]} to provide a more concrete picture. We average the metrics over the number of labels (macro average).

\subsection{Results}
\noindent 
Figure~\ref{fig:carep_eval_grid} reports the performance of our method CAREP against multiple LLM baselines (Claude Sonnet 3.5/3.7, GPT4.1, and GPT4.1 mini). CAREP consistently outperforms all standalone LLMs across both evaluation protocols.

\paragraph{Semantic evaluation} 
In terms of truth-table agreement, CAREP substantially improves over LLMs in capturing the logical structure of error patterns. CAREP achieves a \emph{Recall@1} of $0.70$, compared to $0.25$ for the best-performing baseline (Claude Sonnet 3.7). Precision and F1 scores follow the same trend, showing that CAREP’s causal discovery step yields more faithful Boolean rules rather than over-generalized expressions. Interestingly, LLMs still reach $\approx 0.50$ semantic \emph{Accuracy@5}, indicating that DTCs' observations and descriptions alone allow them to partially approximate the ground truth. However, they lack the consistency and reliability of CAREP, required for deployment.

\paragraph{Structural evaluation}
When evaluating whether the predicted rules include the correct DTCs, the advantage of CAREP becomes even more pronounced. For \emph{Precision@1}, CAREP achieves $0.78$ versus only $0.25$ for Claude Sonnet 3.7 and $0.05$ for GPT4.1 mini. Similar margins are observed in F1 and accuracy. GPT4.1 and mini, in particular, perform poorly (\emph{F1@1} $< 0.1$), demonstrating that without causal discovery, LLMs fail to reason and recover the true DTCs present in the error pattern rules. 

\paragraph{Discussion}
Overall, CAREP exhibits strong gains across all metrics and evaluation modes. The results confirm that pairing a multi-agent system with causal reasoning is critical: while LLMs display non-trivial reasoning abilities in isolation, they remain limited when confronted with high-dimensional, imbalanced rules. A key challenge lies in the cardinality of event types ($|\mathbb{X}|=29{,}100$), which renders direct rule inference from raw sequences infeasible. CAREP addresses this by first extracting the potential causes of each error pattern, thereby purging irrelevant events and selecting only informative candidates. This design allows the orchestrator agent to focus on tractable subsets of events, ultimately producing rules that are both more accurate and interpretable.

\section{Conclusion}
\noindent 
We introduced CAREP, a multi-agent causal reasoning framework that automatizes the discovery of error pattern rules from large-scale event sequences of error codes. By combining causal discovery, RAG system, and agents, CAREP consistently outperforms state-of-the-art LLM baselines while producing interpretable reasoning traces essential for safety-critical deployment. These results position CAREP as a strong step toward practical, trustworthy automation in complex industrial settings.

Beyond automotive applications, we believe CAREP offers an interesting paradigm for agent-based causal reasoning in high-dimensional event-driven systems, with potential extensions to predictive maintenance, robotics, and other domains where reliability and interpretability are paramount.

\newpage

\bibliographystyle{IEEEtran}
\bibliography{math}

\appendix

\begin{table*}[b]
\centering
\caption{\textbf{Ablations of the one-shot phase}. Results are averaged over 5-folds. \emph{Tfy F1} shows the classification performance of \(\text{Tf}_y \) on the test set (multi-label sequence classification), all the other metrics show the structural evaluation performance. \textbf{Bold} means the best result and \uline{Underlined} equally the best. Metrics are given in \(\%\)}
\label{tab:ablation_nades_comparison}
\begin{tabular}{lcccccc}
\toprule
\textbf{Tokens} & \textbf{Parameters} & \textbf{Context} & \textbf{Precision (↑)} & \textbf{Recall (↑)} & \textbf{F1 Score (↑)} & \textbf{Tfy F1 (↑)} \\
\midrule
\multicolumn{7}{c}{\textit{For \(n = 50{,}000\) samples}} \\
\midrule
1.5B & 105m & \(c = 4\)  & \(47.95 \pm 1.05\) & \(30.65 \pm 0.51\) & \(37.39 \pm 0.67\) & 88.6 \\
1.5B & 105m & \(c = 12\) & \(54.62 \pm 1.03\) & \(29.88 \pm 0.73\) & \(38.63 \pm 0.85\) & 90.43 \\
1.5B & 105m & \(c = 15\) & \(\mathbf{55.26 \pm 1.42}\) & \(\uline{31.37} \pm 0.82\) & \(\mathbf{40.02 \pm 1.03}\) & 90.57 \\
1.5B & 105m & \(c = 20\) & \(49.52 \pm 1.59\) & \(\uline{31.76} \pm 0.85\) & \(36.54 \pm 1.10\) & 91.19 \\
1.5B & 105m & \(c = 30\) & \(36.65\pm 1.18\) & \(22.75 \pm 0.78\) & \(26.57 \pm 0.91\) & \textbf{92.64} \\
300m & 47m & \(c = 20\) & \(39.49 \pm 1.77\) & \(26.30 \pm 0.89\) & \(29.01 \pm 1.10\) & 83.6 \\
\bottomrule
\end{tabular}
\end{table*}
\begin{table}[!h]
\caption{Illustrative truth table for semantic evaluation. The accuracy using the predicted rule is \(50\%\).}
\label{tab:truth_table_example}
\centering
\begin{tabular}{|c|c|c|c|}
\hline
\textbf{$x_1$} & \textbf{$x_2$} & \textbf{Ground Truth ($x_1 \; \& \; x_2$)} & \textbf{Predicted Rule ($x_1 \;|\; x_2$)} \\
\hline
0 & 0 & 0 & \textcolor{green}{0} \\
0 & 1 & 0 & \textcolor{red}{1} \\
1 & 0 & 0 & \textcolor{red}{1} \\
1 & 1 & 1 & \textcolor{green}{1} \\
\hline
\end{tabular}
\end{table}
\begin{figure}[!h]
\caption{Example of inputs given to the agentic system for the causal reasoning. The \textit{potential causes} are output by the causal discovery algorithm. For instance, \textit{DTC3} is a cause of \textit{unknown EP} and has an Average Causal Effect (\textit{ACE}) of 0.7, i.e., it increases the likelihood of observing the EP on average by 70\%. Example of defective vehicles with such error pattern are given in \textit{dtcs samples}, metadata are included in it.}
\label{verbatim:output_json}
\centering
\footnotesize 
\begin{lstlisting}[breaklines,frame=single]{json}
"unknown EP": {
  "potential_causes": {
    "DTC3": {
      "frequency": 0.78,
      "ACE_mean": 0.70,
      "ACE_std": 0.13,
      "pmi": {
        "DTC345": -1.89
      }
    },
    "DTC345": {
      "frequency": 0.49,
      "ACE_mean": 0.59,
      "ACE_std": 0.12,
      "pmi": {
        "DTC3": -1.89
      }
    },
    "dtcs_samples": {
      "0X001 0X0008 0X0120 0X8900 .. META .. CCM 111",
      "0x6052 0X0204 0X0129 0X3410 .. META .. CCM 111",
      ...
     }  
    }
}
\end{lstlisting}
\end{figure}
\noindent We present several ablations on the quality of \(\text{Tf}_x, \text{Tf}_y\) on the one-shot causal discovery phase. Table \ref{tab:ablation_nades_comparison} compares the F1, Precision, Recall of the number of trainable \emph{Parameters}, \emph{Context} \(c\), amounts of pretraining data (\emph{Tokens}).
We reported the classification results on the same test set used in the experiments section. The running time was approximately the same for all models: \(11.27\) minutes.
Based on the results, we chose the backbone with 1.5B Tokens, 105M parameters, and a context \(c=15\) for our experiments.


\end{document}